\documentclass[runningheads]{llncs}
\usepackage[T1]{fontenc}
\usepackage{lmodern}
\usepackage{xspace}

\usepackage{graphicx}
\usepackage{hyperref}
\usepackage[misc]{ifsym}
\usepackage{amsmath}
\usepackage{amsfonts}
\usepackage{multirow, makecell}
\usepackage{booktabs}
\usepackage{threeparttable}
\usepackage{microtype}
\usepackage{diagbox}
\usepackage[ruled,vlined,linesnumbered]{algorithm2e}
\usepackage{CJKutf8}

\usepackage{floatrow}
\newfloatcommand{capbtabbox}{table}[][0.42\textwidth]

\ifdefined\directlua
  \usepackage{fontspec}
\else
  \usepackage[T1]{fontenc}
\fi

\DeclareMathOperator*{\argmin}{arg\,min}

\newcommand{\modelname}{ContEA\xspace}
\begin{document}
\title{Facing Changes: Continual Entity Alignment for Growing Knowledge Graphs}

\titlerunning{Continual Entity Alignment for Growing Knowledge Graphs}
%
\author{
  Yuxin Wang\inst{1} \and 
  Yuanning Cui\inst{1} \and 
  Wenqiang Liu\inst{3} \and 
  Zequn Sun\inst{1} \and\\ 
  Yiqiao Jiang\inst{3} \and 
  Kexin Han\inst{3} \and 
  Wei Hu\inst{1,2}\textsuperscript{(\Letter)}
}

\institute{
  State Key Laboratory for Novel Software Technology,\\ 
  Nanjing University, Nanjing, China\\
  \and    
  National Institute of Healthcare Data Science,\\ 
  Nanjing University, Nanjing, China\\
  \and 
  Interactive Entertainment Group, Tencent Inc, Shenzhen, China\\
  \email{yuxinwangcs@outlook.com, \{yncui.nju,zqsun.nju\}@gmail.com} \\
  \email{\{masonqliu,gennyjiang,casseyhan\}@tencent.com, whu@nju.edu.cn}
}

\authorrunning{Y. Wang et al.}
%
\maketitle              
\begin{abstract}
Entity alignment is a basic and vital technique in knowledge graph (KG) integration. 
Over the years, research on entity alignment has resided on the assumption that KGs are static, which neglects the nature of growth of real-world KGs.
As KGs grow, previous alignment results face the need to be revisited while new entity alignment waits to be discovered. 
In this paper, we propose and dive into a realistic yet unexplored setting, 
referred to as continual entity alignment.
To avoid retraining an entire model on the whole KGs whenever new entities and triples come,
we present a continual alignment method for this task.
It reconstructs an entity's representation based on entity adjacency, 
enabling it to generate embeddings for new entities quickly and inductively using their existing neighbors.
It selects and replays partial pre-aligned entity pairs to train only parts of KGs while extracting trustworthy alignment for knowledge augmentation.
As growing KGs inevitably contain non-matchable entities, different from previous works, 
the proposed method employs bidirectional nearest neighbor matching to find new entity alignment and update old alignment.
Furthermore, we also construct new datasets by simulating the growth of multilingual DBpedia.
Extensive experiments demonstrate that our continual alignment method is more effective than baselines based on retraining or inductive learning.

\end{abstract}

\section{Introduction}\label{sect:intro}
Entity alignment, also known as entity matching or entity resolution~\cite{Paris}, has been a long-standing research topic in the Semantic Web and Database communities.
The task aims at matching the identical entities with different URIs
in different knowledge graphs (KGs).
For example, two entities
\nolinkurl{http://dbpedia.org/resource/Hangzhou}
and
\begin{CJK*}{UTF8}{gbsn}\nolinkurl{http://zh.dbpedia.org/resource/}杭州\end{CJK*}
from DBpedia~\cite{DBpedia} in different languages both refer to the same Chinese city, Hangzhou, which is the venue of ISWC 2022 conference.
Early studies \cite{LogMap,Paris} mainly explore the literal similarities with probabilistic or semantic inference to match entities.
However, these methods are hampered by the symbolic heterogeneity of different KGs, particularly the cross-lingual KGs.
To resolve this issue, recent embedding-based methods strive to construct a unified vector space to represent different KGs, 
with entity embeddings used to infer entity similarity~\cite{JAPE}.
Furthermore, 
the embeddings from the unified space built by aligning various KGs are shown to be useful for downstream tasks, 
such as cross-lingual knowledge transfer and multi-lingual KG completion \cite{KGC_EA_EMNLP,KGC_EA_AKBC}.
Thus, as a backbone of knowledge fusion and transfer, 
embedding-based entity alignment has received increasing attention~\cite{OpenEA,EA_survey_AIOpen,TKDE_EA_survey}.

\begin{figure}[!t]
\includegraphics[width=\textwidth]{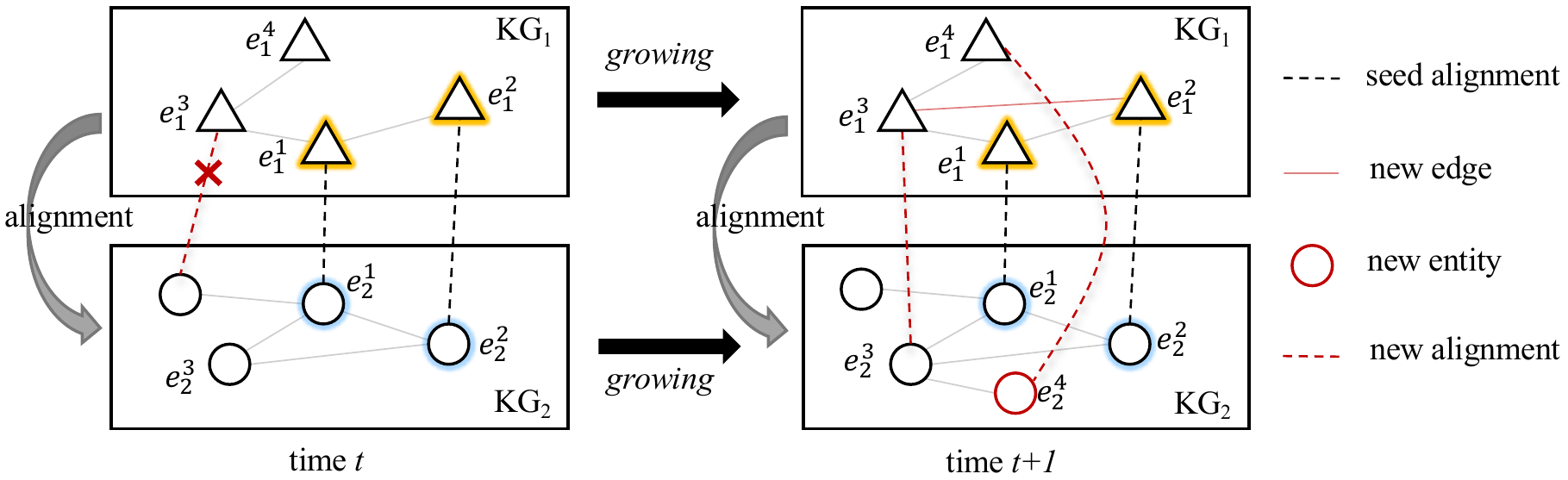}
\caption{Illustration of continual entity alignment. Given two pre-aligned entity pairs $(e_1^1, e_2^1)$ and $(e_1^2, e_2^2)$ between $\text{KG}_1$ and $\text{KG}_2$, we expect to find the identical counterparts for $e_1^3$ and $e_1^4$. At time $t$, due to the incompleteness of both KGs, $e_1^3$ can be falsely matched to a wrong entity, and the expecting counterpart of $e_1^4$ does not even appear yet. At time $t+1$, as new triples emerge over time, $e_1^3$ and $e_1^4$ gain more chance to be correctly matched with richer supportive information.}
\label{fig:example}
\centering
\end{figure}

However, existing embedding-based entity alignment methods assume an idealized scenario of static KGs, neglecting many real-world difficulties like alignment incompleteness, KG growth, and alignment growth. 
In this paper, we argue that entity alignment is not a one-time task. 
We propose and study a new setting, i.e., \emph{continual entity alignment}, between growing and incomplete KGs.
Our motivation comes from the growth and incompleteness nature of real-world KGs.
For example, the release bot of DBpedia \cite{DBpedia} extracts about 21 billion new triples per month \cite{DBpedia_Update}, and Wikidata \cite{Wikidata} releases data dumps in a weekly cycle.\footnote{\url{https://dumps.wikimedia.org/wikidatawiki/entities/}}
The new entities and triples bring about new alignment to be found and provide new clues for correcting the previous alignment.
Fig.~\ref{fig:example} presents an illustration.

This real scenario poses new challenges to embedding-based entity alignment.
The first challenge is \emph{how to learn embeddings for the new entities in an effective and efficient manner}.
When KGs grow, the pre-trained entity alignment model sees new entities for the first time,
as new triples bring structural changes to KGs.
To handle new entities, retraining the model from scratch is costly.
Also, inductive entity embedding is less adaptable to changes of structure.
Thus, it requires non-trivial updates of the pre-trained model to incorporate new entities and new triples.
The second challenge is \emph{how to capture the potential alignment of both old and new entities}.
In real cases, KGs always contain unknown non-matchable entities \cite{DBP2}, which necessitates a more reliable alignment retrieval strategy than simply ranking candidates from test sets.
Furthermore, as new entities typically are few-linked \cite{GEN}, 
capturing the potential alignment for new entities becomes more difficult.
The third challenge is \emph{how to integrate the old predicted alignment with the new predictions}.
In our setting, we output alignment results each time the KGs grow.
The old and new alignment inevitably have conflicts.
We need an effective integration strategy to combine them and update the final alignment.

As the first attempt to address these challenges, we propose a continual entity alignment method \textbf{\modelname}.
Our key idea is to finetune the pre-trained model to incorporate new entities and triples, meanwhile capturing the potential entity alignment.
Specifically, we use Dual-AMN \cite{Dual-AMN}, a prominent alignment model, as our basal encoder.
To enable it to effectively handle new entities, we design an entity reconstruction objective, which allows the encoder to generate entity embeddings using solely neighboring subgraphs.
To retrieve alignment from the embedding space,
we propose a bidirectional nearest neighbor search strategy.
Two entities are predicted to be aligned if and only if they are the nearest neighbors to each other.
When new entities and triples emerge, 
\modelname finetunes the pre-trained model according to the changed structures.
To capture potential entity alignment,
we replay partial pre-known alignment to avoid knowledge oblivion and select high-confidence predictions for knowledge augmentation.

To support the research on this new and practical task, 
we build three new datasets based on the widely-used benchmark DBP15K~\cite{JAPE}, which contains three cross-lingual datasets, i.e., ZH-EN, JA-EN and FR-EN.
For each dataset, we construct six snapshots (i.e., $t=0, 1,2,3,4,5$) by adding new entities and new triples into the preceding snapshot, to simulate KGs' growth. 
We conduct extensive experiments on our datasets.
Our method outperforms strong baselines that use retraining or inductive embedding techniques while at a lower time cost.
Our datasets and source code are publicly available to foster future research.

\section{Problem Statement}\label{sect:task}
We define a KG as a 3-tuple $\mathcal{G}=\{\mathcal{E}, \mathcal{R}, \mathcal{T}\}$, where $\mathcal{E}$ and $\mathcal{R}$ denote the sets of entities and relations, respectively. $\mathcal{T} \subseteq \mathcal{E} \times \mathcal{R} \times \mathcal{E}$ is the set of relational triples.
Given two KGs $\mathcal{G}_1=\{\mathcal{E}_1, \mathcal{R}_1, \mathcal{T}_1\}$ and $\mathcal{G}_2=\{\mathcal{E}_2, \mathcal{R}_2, \mathcal{T}_2\}$, \emph{entity alignment} aims to identify entities in $\mathcal{G}_1$ and $\mathcal{G}_2$ that refer to the same real-world object, 
i.e., seeking a set of alignment $\mathcal{A} = \{(e_1, e_2) \in \mathcal{E}_s \times \mathcal{E}_t \,|\, e_1 \equiv e_2\}$,
where ``$\equiv$'' indicates equivalence.
A small set of seed entity alignment $\mathcal{A}_{\text{s}}\subset\mathcal{A}$ is usually provided as anchors (i.e., training data) beforehand to help align the remaining entities.

From time to time, new triples emerge and are added into KGs, which brings KGs' size growth. 
We propose the definition of \textit{growing KGs} as follows: 

\begin{definition}[Growing knowledge graphs] 
A growing KG $\mathcal{G}$ is a sequence of snapshots $\mathcal{G} = (\mathcal{G}^0, \mathcal{G}^1, \dots, \mathcal{G}^{T})$, where the superscript numbers denote different timestamps. For any two successive timestamps $\mathcal{G}^{t} = \{\mathcal{E}^t, \mathcal{R}^t, \mathcal{T}^t\}$ and $\mathcal{G}^{t+1} = \{\mathcal{E}^{t+1}, \mathcal{R}^{t+1}, \mathcal{T}^{t+1}\}$, there exist $\mathcal{E}^{t} \subseteq \mathcal{E}^{t+1}$, $\mathcal{R}^{t} = \mathcal{R}^{t+1}$ and $\mathcal{T}^{t} \subseteq \mathcal{T}^{t+1}$. 
\end{definition}

In this definition, each newly added triple in $\rm{\Delta} \mathcal{T}^{t+1}$ between $t$ and $t+1$ contains zero, one, or two new entities. Considering that the set of relations in KGs is much less diverse than that of entities, we dismiss the emergence of new relations in this paper and assume that the relations in KGs are pre-defined. 

To practice entity alignment on growing KGs.
We propose the task of \textit{continual entity alignment} and give its definition below: 

\begin{definition}[Continual entity alignment] Given two growing KGs $\mathcal{G}_{1}$ and $\mathcal{G}_{2}$, and the seed entity alignment $\mathcal{A}_{\text{s}}$ at time $t=0$, continual entity alignment at time $t$ aims to find potential entity alignment $\mathcal{A}_{\text{p}}^{t}$ between $\mathcal{G}_{1}^{t}$ and $\mathcal{G}_{2}^{t}$ based on the currently learned KG embeddings and alignment model. 
\end{definition}

In this definition, the size of $\mathcal{A}_{\text{s}}$ is constant, while $\mathcal{A}_{\text{p}}^{t}$ grows over time as new entities may bring new entity alignment to be found. 
Considering that the seed entity alignment is usually deficient and difficult to obtain \cite{SensitiveLP}, 
we do not assume that new snapshots bring new seed alignment to augment training data.
That is to say, $\mathcal{A}_{\text{s}}$ of snapshot at time $t>0$ is the same as that at time $t=0$.

\section{Methodology}\label{sect:method}

\begin{figure}[!t]
\includegraphics[width=\textwidth]{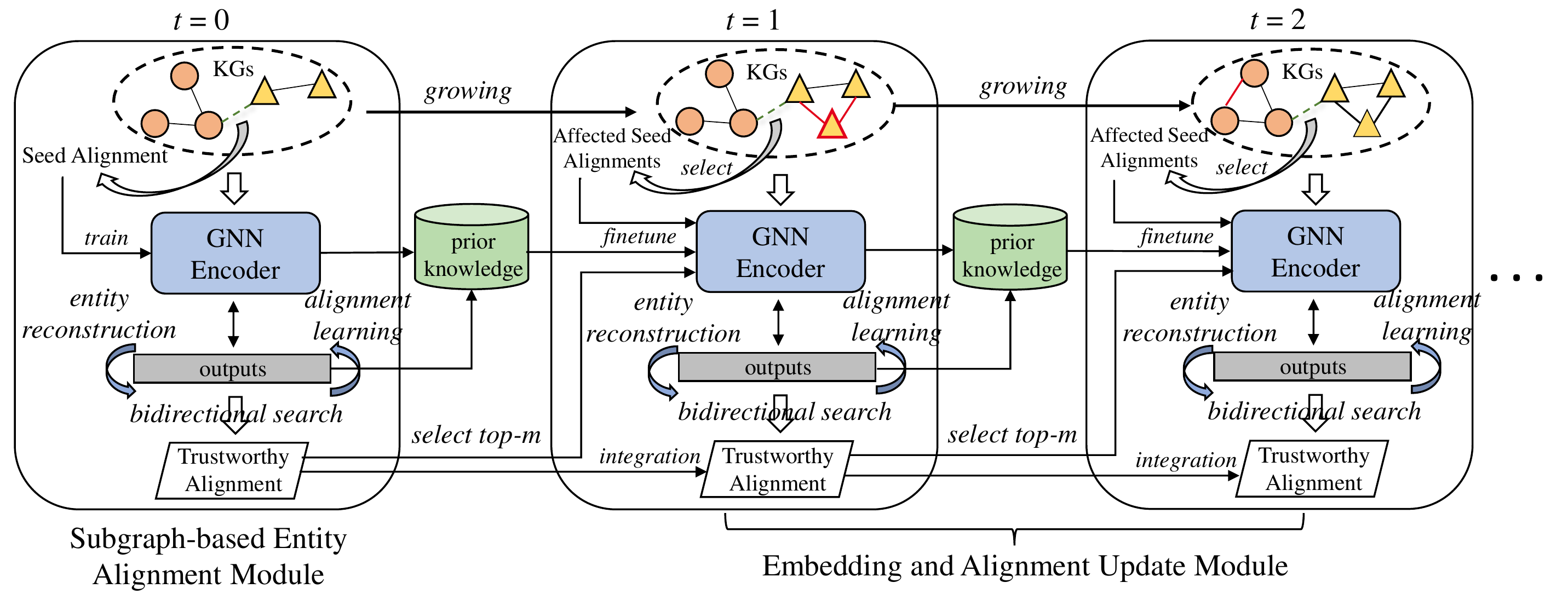}
\caption{Framework of the proposed continual entity alignment method {\modelname}.}
\label{fig:framework}
\centering
\end{figure}

In this section, we introduce the proposed continual entity alignment method \modelname. 
Fig.~\ref{fig:framework} depicts its framework.
It consists of two modules: the subgraph-based entity alignment module, and the embedding and alignment update module. 
The following is a brief overview of them:

\begin{itemize}

\item In the subgraph-based entity alignment module, the input is the two KGs at time $t=0$ and the seed entity alignment across them.
A graph neural network (GNN) is employed over the two KGs to represent entities based on their subgraph structures.
The alignment learning objective is to minimize the embedding distance of similar entities while separating dissimilar ones.
Additionally, an entity reconstruction design is used to encourage entities similar to their contexts.
When the learning process is completed, 
the trustworthy alignment is predicted based on bidirectional nearest neighbor search.

\smallskip
\item At time $t>0$, the embedding and alignment update module first incorporates new entities into previously learned KG embeddings.  
It reconstructs new entities' embeddings based on their neighborhood subgraphs. 
Then, partial seed entity alignment and trustworthy alignment predicted in the previous snapshot are used for finetuning the GNN model.
Last, after new alignment is predicted, we use it to update the previously-found old alignment. 
\end{itemize}

We introduce the details of the two modules in the following two subsections.

\subsection{Subgraph-based Entity Alignment}
This module is built upon a GNN that represents an entity by aggregating its neighborhood subgraph.
The key assumption behind GNN is that the entities with similar neighborhoods appear to be close, 
which makes GNNs extensible to represent new entities.
Please note that we do not focus on how to develop a powerful GNN for entity alignment,
but on how to incorporate new entities and triples in an effective and efficient manner for continual entity alignment.

\subsubsection{Subgraph encoder.}
We adopt the GNN-based encoder of Dual-AMN \cite{Dual-AMN} as our subgraph encoder for its effectiveness and simplicity.
The encoder of Dual-AMN consists of an inner-graph layer (namely $\texttt{Aggregator}_1$) capturing the structural information within a
single KG, and a cross-graph layer ($\texttt{Aggregator}_2$) capturing cross-graph matching information based on the outputs of $\texttt{Aggregator}_1$. 
Technically, $\texttt{Aggregator}_1$ is a 2-layered relation-aware GNN, and $\texttt{Aggregator}_2$ is a proxy attention network connecting entities with a list of proxy nodes.
Overall, given an entity $e$, its representation after being encoded by Dual-AMN is
\begin{equation}
    \texttt{Encoder}(e) = \texttt{Aggregator}_2\big(\texttt{Aggregator}_1(e, \mathcal{N}_e), \mathcal{E}_{\text{proxy}}\big),
\end{equation}
where $\texttt{Aggregator}_1()$ aggregates the entity itself and its relational neighbors $\mathcal{N}_e$ to generate its embedding, 
and $\texttt{Aggregator}_2()$ combines the output embeddings with proxy nodes $\mathcal{E}_{\text{proxy}}$ to generate the final representations of entities.
To save space, we do not present the detailed techniques of Dual-AMN here.
Interested readers can refer to its original paper \cite{Dual-AMN} for more details.

\subsubsection{Entity reconstruction.}
As KGs grow, the pre-trained GNN encoder encounters new entities and triples.
The critical challenge is how to incorporate unseen entities into the encoder.
Randomly initializing the embeddings of new entities could be detrimental to the previously optimized embedding space and cause representation inconsistency.
A typical assumption in embedding-based entity alignment is that two entities are similar if their neighborhood subgraphs are similar (i.e., the two subgraphs have similar or pre-aligned entities).
Motivated by this, we propose a self-supervised learning objective that enables the encoder to reconstruct an entity using its neighborhood subgraphs:
\begin{equation}
    \mathcal{L}_{\text{reconstruct}} = \sum_{e \in \mathcal{E}} \left\Vert\mathbf{e} - \frac{1}{|\mathcal{N}_{e}|} \sum_{e' \in \mathcal{N}_e} \mathbf{e}' \right\Vert^{2}_{2}.
\end{equation}

Here, $\mathcal{N}_e$ denotes the set of one-hop neighbors of $e$. This objective minimizes the distance between an entity and its neighbor subgraph embedding (the mean vector of all neighbor embeddings).

\subsubsection{Alignment learning.}
Given the outputs of the encoder, 
alignment learning aims to gather similar entity pairs and distance dissimilar entity pairs. 
The dissimilar entity pairs is modeled by negative sampling. 
Following Dual-AMN~\cite{Dual-AMN}, we also adopt the \texttt{LogSumExp} function to compute the loss:
\begin{equation}\label{eq:3}
    \mathcal{L}_{\text{align}} = \log \Big[1 + \sum_{(e_1,e_2)\in \mathcal{A}_{\text{s}}} \sum_{(e_1,e_2')\in \mathcal{A}^{\text{neg}}_{e_1}} \exp \big(\gamma(\lambda + \texttt{sim}(e_1,e_2) - \texttt{sim}(e_1,e_2'))
    \big)\Big],
\end{equation}
where $\mathcal{A}^{\text{neg}}_{e_1}$ denotes the negative alignment generated for entity $e_1$. $\gamma$ is a scale factor, and $\lambda$ is the margin for separating the similarities of seed alignment pairs and negative pairs.
\texttt{Cosine} is used to compute embedding similarity, i.e., $\texttt{sim}(e_1,e_2)=\cos(\texttt{Encoder}(e_1), \texttt{Encoder}(e_2))$.
We employ the in-batch negative generating method.
Specifically, for entity $e_1$, other entities (e.g., $e_2'$) in a training batch act as its negative counterparts to generate the negative pairs $\mathcal{A}^{\text{neg}}_{e_1}$. The final learning objective of subgraph-based entity alignment module $\mathcal{L}_{1}$ is a combination of $\mathcal{L}_{\text{align}}$ and $\mathcal{L}_{\text{reconstruct}}$ with a weight $\alpha$ on $\mathcal{L}_{\text{reconstruct}}$: 
\begin{equation}\label{eq:4}
    \mathcal{L}_{1} = \mathcal{L}_{\text{align}} + \alpha \cdot \mathcal{L}_{\text{reconstruct}}.
\end{equation}

\subsubsection{Trustworthy alignment search.}\label{alignment_search}
After the alignment learning is complete,
we retrieve trustworthy entity alignment as predictions based on the optimized embedding space.
Previous embedding-based entity alignment methods assume that each entity in one KG must have a counterpart in the other KG.
A typical inference process is the nearest neighbor search, i.e., it seeks
\begin{equation} 
\label{eq:inference}
\hat{e}_2=\argmin_{e_2\in \mathcal{E}_2} \pi(\texttt{Encoder}(e_1), \texttt{Encoder}(e_2)),
\end{equation}
where $\pi()$ is a measure for alignment search, and $\hat{e}_2$ is the predicted counterpart for $e_1$.
However, such an ``idealized'' assumption may not stand in a realistic setting as there are many no-match entities in the two KGs \cite{DBP2}.
To resolve this issue and improve alignment search,
we propose a parameter-free strategy called bidirectional nearest alignment search.
It searches for the nearest neighbor in one KG for the entities in the other.
An alignment pair $(e_1, e_2)$ is a trustworthy alignment
if and only if $e_2=\hat{e}_2$ and $e_1=\hat{e}_1$.
Other alignment pairs are discarded.

\subsection{Embedding and Alignment Update}
At time $t>0$, the relational structure of KGs get changed as new triples come.
It needs to generate embeddings for new entities while capturing the structure changes.
To resolve this challenge, we propose to \textit{finetune} the GNN encoder and new entity embeddings with partial seed alignment and selected trustworthy alignment.
After finetuning, the new trustworthy entity alignment is retrieved based on the updated model and embeddings.
The new predicted alignment is used to complete and update the old alignment discovered at time $t-1$ using a heuristic strategy.

\subsubsection{Encoder finetuning.}
We initialize the encoder with the parameters learned in the previous module/time. 
Thanks to our entity reconstruction objective, the encoder is able to initialize the embedding of a new entity $e$ as follows:
\begin{equation}\label{eq:new_copy}
    \texttt{Encoder}(e) =  \texttt{Aggregator}_2\big(\texttt{Aggregator}_1(\texttt{MP}(\mathcal{N}'_e)), \mathcal{E}_{\text{proxy}}\big),
\end{equation}
where $\mathcal{N}'_e$ denotes the seen neighbors of the new entity $e$. $\texttt{MP}()$ is mean-pooling process to generate embedding for $e$ using $\mathcal{N}'_e$.

Based on the output embeddings of new and existing entities, we finetune the GNN encoder.
Specifically, we freeze the inner-graph layer $\texttt{Aggregator}_1$ while make the cross-graph $\texttt{Aggregator}_2$ learnable. 
For a single KG, the coming of new data does not change the neighbor aggregation pattern, as a KG’s schema stays consistent (no new relations or entity domains). But the two KGs grow independently and asymmetrically in the proposed scenario. 
It is necessary to fine-tune the matching network to make adjustments and new discoveries.

For training data, considering that the potential entity alignment is more likely to occur near anchors \cite{iMAP},
we replay only the affected seed entity alignment that contains anchors involved in new triples.
This helps the alignment of new entities, which is originally difficult due to their low degrees.
Also, to help align entities from wider and more dynamic areas, we select top-$m$ predicted trustworthy alignment with the highest similarity scores and treat them as ``new anchors''. 

We finetune the GNN encoder and new entity embeddings on the obtained affected seed alignment (ASA for short) and $m$ selected trustworthy alignment (TA for short).
We use a weight $\beta$ on the learning loss over $m$ trustworthy alignment to balance its importance. 
The final loss function $\mathcal{L}_2$ of finetuning is
\begin{equation}\label{eq:7}
    \mathcal{L}_{\text{2}} = \mathcal{L}_{\text{align}}(ASA) + \alpha \cdot \mathcal{L}_{\text{reconstruct}} + \beta \cdot \mathcal{L}_{\text{align}}(TA).
\end{equation}

\subsubsection{Trustworthy alignment update.} \label{alignment_update}
After finetuning, a new set of trustworthy alignment can be retrieved using the updated entity embeddings and model.
It is necessary to combine it with the previously discovered trustworthy alignment because they are gathered from different snapshots and may complement each other to produce superior outcomes.
Here, we carry out a heuristic strategy to integrate them.
We keep new trustworthy alignment which is between two new entities.
But for new ones that cause alignment conflicts \cite{BootEA} with the previous trustworthy alignment (i.e., an entity is aligned with different entities), we decide to keep the alignment that has higher similarity scores.
With KGs growing, the size of trustworthy entity alignment is accumulative. 

\subsection{Put It All Together}
Algorithm~\ref{alg:ContEA} describes the training and finetuning details of \modelname for continual entity alignment.
Lines 1--3 describe the process of the subgraph-based entity alignment module at time $t=0$. 
Lines 4--8 describe the process of embedding and alignment updating modules at time $t>0$.

\begin{algorithm}[!tb]
\caption{Process of \modelname}
\label{alg:ContEA}
\SetKwInOut{Input}{Input}\SetKwInOut{Output}{Output}
\Input{Two growing KGs $G_1^t$ and $G_2^t$ at time $t$, prior learned knowledge $\mathbf{K}$ (none for $t=0$), seed alignment $\mathcal{A}_{\text{s}}$, previous trustworthy alignment $TA$ (none for $t=0$), hyperparameters $\alpha$, $\beta$;}
\Output{Updated trustworthy alignment $TA$;}
 \eIf{$t=0$}{
  Training encoder on $G_1^0$ and $G_2^0$ using $\mathcal{L}_{1}$ loss in Eq.~(\ref{eq:4})\;
  Generating $TA$ though trustworthy alignment search\;
 }
 {
  Initializing embeddings and encoder parameters using $\mathbf{K}$, $G_1^t$ and $G_2^t$\;
  Selecting affected $\mathcal{A}_{\text{s}}$ as $ASA$\ and top-$m$ $TA$ with highest similarity\;
  Finetuning encoder using $\mathcal{L}_{2}$ loss in Eq.~(\ref{eq:7})\;
  Updating $TA$ with new trustworthy alignment\;
 }
\end{algorithm}
\section{Experiments}

\subsection{New Datasets for Continual Entity Alignment}
Due to the lack of off-the-shelf benchmarks for proposed setting, we construct new datasets based on DBP15K \cite{JAPE}.
For each DBP15K's cross-lingual entity alignment dataset, we use its two KGs as the first snapshots (i.e., $t=0$).
DBP15K only considers entity alignment between the head entities of triples and overlooks other entity alignment pairs.
Hence, we first complete the reference entity alignment using the inter-language links in DBpedia\footnote{We use the infobox-based relation triples (version 2016-10) following DBP15K.}, resulting in more than 15K reference alignment pairs in the first snapshot.
Then, the reference entity alignment is divided into training, validation and test sets (i.e., $\mathcal{A}_\text{s}$, $\mathcal{A}_\text{v}$ and $\mathcal{A}_\text{p}^{0}$) with a ratio of $2:1:7$. 
We further build five snapshots to simulate KGs' growth: 
\begin{itemize}
    \item At time $t>0$, we first collect the relation triples from DBpedia that contain entities in $\mathcal{G}_1^{t-1}$ and $\mathcal{G}_2^{t-1}$.
    Then, among these triples we remove seen ones at time $t-1$, and sample new triples from the remaining with the size of $20\%$ of the triples in previous snapshots. Adding the new triples into $\mathcal{G}_1^{t-1}$ and $\mathcal{G}_2^{t-1}$ and we create snapshots $\mathcal{G}_1^{t}$ and $\mathcal{G}_2^{t}$.
    
    \smallskip\item Then, we complete $\mathcal{G}_1^{t}$ and $\mathcal{G}_2^{t}$ by adding additional relation triples from DBpedia of which the head and tail entities are both in the snapshots, leading to more than $20\%$ growth of triple size.
    
    \smallskip\item Finally, we retrieve the new entity alignment pairs brought by the newly added entities, and add them into the test set $\mathcal{A}_\text{p}^{t}$ of snapshot $t$.
    The training set $\mathcal{A}_\text{s}$ or validation set $\mathcal{A}_\text{v}$ still follows that in the first snapshot at time $t=0$.
    We do not assume that the new snapshot introduces new training data because obtaining seed alignment for emerging entities is usually more difficult than finding seed alignment for old entities in the real world.

\end{itemize}

The detailed statistics of our dataset are present in Table~\ref{table: datasets}.
\begin{table*}[!tb]\setlength\tabcolsep{2pt}
\centering
\caption{Statistics of the three datasets. Each consists of two growing KGs in six snapshots from consecutive timestamps. In a snapshot, $|\mathcal{T}|$ is the current triple size, and $|\mathcal{A}_{\text{s}}|, |\mathcal{A}_{\text{v}}|, |\mathcal{A}_{\text{p}}|$ are the sizes of training, validation and test alignment, respectively.}
\label{table: datasets}
\resizebox{\textwidth}{!}{
\begin{tabular}{c|ccccc|ccccc|ccccc}
  \toprule
   & \multicolumn{5}{c|}{$\text{DBP}_\text{ZH-EN}$} 
   & \multicolumn{5}{c|}{$\text{DBP}_\text{JA-EN}$} 
   & \multicolumn{5}{c}{$\text{DBP}_\text{FR-EN}$} \\
  \cmidrule(lr){2-6} \cmidrule(lr){7-11} \cmidrule(lr){12-16}
  & $|\mathcal{T}|_{\text{ZH}}$ & $|\mathcal{T}|_{\text{EN}}$  & $|\mathcal{A}_{\text{s}}|$ & $|\mathcal{A}_{\text{v}}|$ & $|\mathcal{A}_{\text{p}}|$ & $|\mathcal{T}|_{\text{JA}}$ & $|\mathcal{T}|_{\text{EN}}$  & $|\mathcal{A}_{\text{s}}|$ & $|\mathcal{A}_{\text{v}}|$ & $|\mathcal{A}_{\text{p}}|$ & $|\mathcal{T}|_{\text{FR}}$ & $|\mathcal{T}|_{\text{EN}}$  & $|\mathcal{A}_{\text{s}}|$ & $|\mathcal{A}_{\text{v}}|$ & $|\mathcal{A}_{\text{p}}|$ \\
  \midrule
  $t=0$ & \ \,70,414 & \ \,95,142 & 3,623 & 1,811 & 12,682 & \ \,77,214 & \ \,93,484 & 3,750 & 1,875 & 13,127 & 105,998 & 115,722 & 3,727 & 1,863 & 13,048  \\
  $t=1$ & 103,982 & 154,833 & 3,623 & 1,811 & 14,213 & 112,268 & 150,636 & 3,750 & 1,875 & 15,079 & 148,274 & 184,132 & 3,727 & 1,863 & 15,875  \\
  $t=2$ & 137,280 & 213,405 & 3,623 & 1,811 & 16,296 & 147,097 & 207,056 & 3,750 & 1,875 & 18,092 & 191,697 & 251,591 & 3,727 & 1,863 & 20,481 \\
  $t=3$ & 173,740 & 278,076 & 3,623 & 1,811 & 18,716 & 185,398 & 270,469 & 3,750 & 1,875 & 21,690 & 239,861 & 326,689 & 3,727 & 1,863 & 25,753 \\
  $t=4$ & 213,814 & 351,659 & 3,623 & 1,811 & 21,473 & 227,852 & 341,432 & 3,750 & 1,875 & 25,656 & 293,376 & 411,528 & 3,727 & 1,863 & 31,564 \\
  $t=5$ & 258,311 & 434,683 & 3,623 & 1,811 & 24,678 & 274,884 & 421,971 & 3,750 & 1,875 & 29,782 & 352,886 & 507,793 & 3,727 & 1,863 & 37,592 \\
  \bottomrule
\end{tabular}}
\end{table*}

\subsection{Baselines}
We compare \modelname with two groups of entity alignment methods. 
\begin{itemize}
    \item \textbf{Retraining baselines.} Since most existing embedding-based EA methods are designed for static KGs, they need retraining each time new triples come. 
    Here, we choose the representative translation-based method MTransE \cite{MTransE}, and several state-of-the-art GNN-based methods, including GCN-Align \cite{GCN-Align}, AlignE \cite{BootEA}, AliNet \cite{AliNet}, KEGCN \cite{KEGCN} and Dual-AMN \cite{Dual-AMN} as our baselines.
    
    \smallskip\item \textbf{Inductive baselines.} 
    The only entity alignment method focusing on KGs' growth is DINGAL \cite{DINGAL}. We choose one of the proposed variants, DINGAL-O, as a baseline, which can handle our scenario. 
    Additionally, since there are some inductive KG embedding (KGE) methods which can generate embeddings for new entities, we explore their combination with static methods to tackle our task. 
    Here, we select two representative inductive KGE methods MEAN \cite{MEAN} and LAN \cite{LAN} as the entity representation layer and incorporate them with Dual-AMN. 
    We denote the two baselines by $\rm{MEAN}^+$ and $\rm{LAN}^+$.
\end{itemize}

\subsection{Experiment Settings}
\subsubsection{Evaluation metrics.}
At each time $t$, the bidirectional nearest neighbor search and alignment integration are used to obtain the final trustworthy alignment. 
The details are described in Sect.~\ref{alignment_search} and Sect.~\ref{alignment_update}.
Then, we compare the final trustworthy alignment with gold test pairs $\mathcal{A}_{\text{p}}^{t}$.
We report the precision, recall, and F1 scores as the evaluation metrics.

\subsubsection{Implementation.}
We implement \modelname, Dual-AMN, $\rm{MEAN}^+$ and $\rm{LAN}^+$ using PyTorch.
For other retraining baselines, we use the implementations in an open-source library.\footnote{\url{https://github.com/nju-websoft/OpenEA}}
We set the embedding dimensions to 100.
The embedding similarity metric is CSLS \cite{CSLS}.
We use grid search on hyperparameters and early stop to find the best performance.
Specifically for \modelname, we set $\alpha=0.1$, $\beta=0.1$ and $m=500$.
More detailed hyperparameter settings can be found on our GitHub repository.
For a fair comparison, all baselines only rely on KGs' structural information and do not use pre-trained models for initialization.

\subsection{Results}

\subsubsection{General results.}
We conduct experiments on the constructed datasets and present the results in Tables~\ref{tab:zh_main_results}, \ref{tab:ja_main_results} and \ref{tab:fr_main_results}.
Compared with baselines, \modelname reaches the best performance in discovering potential entity alignment. 
Its F1 scores outperform the best baseline Dual-AMN by 27.1\%, 19.4\%, and 15.2\% averagely on six snapshots of $\text{DBP}_\text{ZH-EN}$, $\text{DBP}_\text{JA-EN}$, and $\text{DBP}_\text{FR-EN}$, respectively.
The superior performance of \modelname over retraining methods is because \modelname can iteratively leverage the prior knowledge (e.g., previously predicted alignment and model parameters) from the past snapshots. 
Also, \modelname collectively obtains predicted entity alignment by integrating new and old trustworthy alignment rather than totally neglecting old predictions in retraining.
As for inductive baselines, $\rm{MEAN}^{+}$ and $\rm{LAN}^{+}$ perform worse than \modelname and Dual-AMN, which indicates that straightway adding the inductive KGE layer without adjusting the alignment network does not give satisfactory performance.
DINGAL-O also shows unsatisfactory results, because it is purely inductive and does not update the alignment network.
Besides, we can notice that the performance of all methods declines over time. 
This is due to the expansion of the searching space for alignment candidates, and the drop in the ratio of seed alignment against to-be-aligned alignment.
Both of these increase the probability of entities being mismatched. 

\begin{table*}[!tb]\setlength\tabcolsep{2pt}
\centering
\caption{Results of entity alignment on $\text{DBP}_\text{ZH-EN}$. NA stands for Not Applicable.}
\label{tab:zh_main_results}
\resizebox{\textwidth}{!}{
    \begin{tabular}{cl|cccccc}
    \toprule
        & & $t=0$ & $t=1$ & $t=2$ & $t=3$ & $t=4$ & $t=5$ \\
        \cmidrule(lr){3-3} \cmidrule(lr){4-4} \cmidrule(lr){5-5} \cmidrule(lr){6-6} \cmidrule(lr){7-7} \cmidrule(lr){8-8}
         & & P\space/ R /\space F1 & P\space/ R /\space F1 & P\space/ R /\space F1 & P\space/ R /\space F1 & P\space/ R /\space F1 & P\space/ R /\space F1 \\
    \midrule
        \parbox[t]{3mm}{\multirow{6}{*}{\rotatebox[origin=c]{90}{Retraining}}} 
        & MTransE & .552/.178/.269 & .242/.111/.152 & .159/.078/.105 & .094/.054/.068 & .080/.041/.055 & .049/.030/.037 \\
        & GCN-Align & .550/.249/.343 & .212/.152/.177 & .133/.115/.123 & .096/.091/.094 & .076/.075/.076 & .062/.062/.062 \\
        & AlignE & .721/.364/.484 & .382/.272/.317 & .282/.222/.248 & .206/.173/.188 & .191/.152/.169 & .127/.112/.119 \\
        & AliNet & .641/.358/.459 & .285/.311/.297 & .195/.279/.230 & .146/.244/.183 & .129/.232/.166 & .105/.199/.128 \\
        & KEGCN & .664/.200/.308 & .315/.129/.183 & .198/.093/.127 & .160/.075/.102 & .136/.064/.087 & .120/.052/.072 \\
        & Dual-AMN & .834/.596/.695 & .482/.443/.462 & .357/.356/.356 & .285/.286/.286 & .249/.254/.251 & .227/.227/.227 \\
    \midrule
        \parbox[t]{3mm}{\multirow{3}{*}{\rotatebox[origin=c]{90}{Induct.}}} 
        & $\text{MEAN}^+$ & .828/.576/.679 & .483/.422/.450 & .357/.341/.349 & .267/.264/.265 & .225/.226/.225 & .198/.197/.198 \\
        & $\text{LAN}^+$ & .827/.576/.679 & .488/.426/.455 & .360/.345/.352 & .274/.271/.272 & .231/.229/.230 & .205/.199/.202 \\
        & DINGAL-O & .497/.195/.280 & .370/.158/.222 & .315/.135/.189 & .251/.111/.154 & .229/.093/.132 & .209/.080/.116 \\
    \midrule
        \multicolumn{2}{l|}{\modelname} & \textbf{.843}/\textbf{.604}/\textbf{.703} & \textbf{.555}/\textbf{.539}/\textbf{.546} & \textbf{.444}/\textbf{.473}/\textbf{.458} & \textbf{.373}/\textbf{.421}/\textbf{.396} & \textbf{.324}/\textbf{.375}/\textbf{.348} & \textbf{.291}/\textbf{.336}/\textbf{.312} \\
        \multicolumn{2}{l|}{w/o TA} & NA / NA / NA & .543/.531/.537 &.419/.469/.443 & .357/.414/.384 & .316/.371/.341 & .286/.332/.307 \\
        \multicolumn{2}{l|}{w/o TA \& ASA} & NA / NA / NA & .543/.527/.535 & .422/.463/.442 & .352/.410/.379 & .309/.365/.335 & .278/.324/.300 \\
        \multicolumn{2}{l|}{Retraining} & NA / NA / NA & .493/.455/.473 & .364/.357/.361 & .300/.301/.301 & .265/.266/.265 & .245/.240/.243 \\
    \bottomrule
    \end{tabular}}
\end{table*}
\begin{table*}[!tb]\setlength\tabcolsep{2pt}
\centering
\caption{Results of entity alignment on $\text{DBP}_\text{JA-EN}$. NA stands for Not Applicable.}
\label{tab:ja_main_results}
\resizebox{\textwidth}{!}{
    \begin{tabular}{cl|cccccc}
    \toprule
        & & $t=0$ & $t=1$ & $t=2$ & $t=3$ & $t=4$ & $t=5$ \\
        \cmidrule(lr){3-3} \cmidrule(lr){4-4} \cmidrule(lr){5-5} \cmidrule(lr){6-6} \cmidrule(lr){7-7} \cmidrule(lr){8-8}
        & & P\space/ R /\space F1 & P\space/ R /\space F1 & P\space/ R /\space F1 & P\space/ R /\space F1 & P\space/ R /\space F1 & P\space/ R /\space F1 \\
    \midrule
        \parbox[t]{3mm}{\multirow{6}{*}{\rotatebox[origin=c]{90}{Retraining}}} 
        & MTransE & .599/.200/.299 & .293/.121/.172 & .213/.082/.118 & .151/.061/.087 & .128/.046/.067 & .117/.035/.054 \\
        & GCN-Align & .594/.279/.379 & .263/.183/.216 & .177/.142/.158 & .140/.117/.127 & .116/.099/.107 & .099/.084/.091 \\
        & AlignE & .738/.359/.483 & .433/.282/.342 & .320/.218/.260 & .270/.178/.214 & .228/.148/.180 & .193/.122/.149 \\
        & AliNet & .661/.364/.469 & .305/.312/.308 & .216/.270/.240 & .167/.231/.194 & .149/.215/.176 & .126/.189/.151 \\
        & KEGCN & .663/.198/.305 & .389/.153/.219 & .280/.110/.157 & .245/.087/.128 & .200/.070/.104 & .194/.063/.096 \\
        & Dual-AMN & \textbf{.861}/.606/.711 & .517/.437/.474 & .398/.347/.370 & .348/.292/.318 & .313/.251/.278 & .300/.231/.261 \\
    \midrule
        \parbox[t]{3mm}{\multirow{3}{*}{\rotatebox[origin=c]{90}{Induct.}}} 
        & $\text{MEAN}^+$ & .847/.571/.682 & .528/.420/.468 & .407/.330/.365 & .330/.261/.292 & .287/.221/.250 & .265/.193/.223 \\
        & $\text{LAN}^+$ & .845/.575/.684 & .528/.424/.470 & .410/.333/.368 & .335/.265/.296 & .296/.226/.257 & .274/.200/.231 \\
        & DINGAL-O & .540/.227/.320 & .391/.174/.241 & .328/.137/.194 & .271/.113/.159 & .249/.092/.134 & .231/.078/.116 \\
    \midrule
        \multicolumn{2}{l|}{\modelname} & .858/\textbf{.610}/\textbf{.713} & \textbf{.586}/\textbf{.519}/\textbf{.551} & \textbf{.483}/\textbf{.440}/\textbf{.461} & \textbf{.417}/\textbf{.381}/\textbf{.398} & \textbf{.375}/\textbf{.336}/\textbf{.354} & \textbf{.344}/\textbf{.299}/\textbf{.320} \\
        \multicolumn{2}{l|}{w/o TA} & NA / NA / NA & .572/.518/.544 &.466/.439/.452 & .398/.377/.387 & .357/.332/.344 & .333/.294/.312 \\
        \multicolumn{2}{l|}{w/o TA \& ASA} & NA / NA / NA & .580/.514/.545 & .466/.436/.450 & .399/.374/.386 & .359/.328/.343 & .331/.291/.310 \\
        \multicolumn{2}{l|}{Retraining} & NA / NA / NA  & .530/.449/.486 & .415/.356/.383 & .369/.298/.330 & .349/.272/.306 & .327/.244/.280 \\
    \bottomrule
    \end{tabular}}
\end{table*}
\begin{table*}[!tb]\setlength\tabcolsep{2pt}
\centering
\caption{Results of entity alignment on $\text{DBP}_\text{FR-EN}$. NA stands for Not Applicable.}
\label{tab:fr_main_results}
\resizebox{\textwidth}{!}{
    \begin{tabular}{cl|cccccc}
    \toprule
        & & $t=0$ & $t=1$ & $t=2$ & $t=3$ & $t=4$ & $t=5$ \\
        \cmidrule(lr){3-3} \cmidrule(lr){4-4} \cmidrule(lr){5-5} \cmidrule(lr){6-6} \cmidrule(lr){7-7} \cmidrule(lr){8-8}
        & & P\space/ R /\space F1 & P\space/ R /\space F1 & P\space/ R /\space F1 & P\space/ R /\space F1 & P\space/ R /\space F1 & P\space/ R /\space F1 \\
    \midrule
        \parbox[t]{3mm}{\multirow{6}{*}{\rotatebox[origin=c]{90}{Retraining}}} 
        & MTransE & .570/.188/.283 & .246/.100/.142 & .145/.062/.087 & .108/.040/.059 & .104/.032/.049 & .073/.024/.036 \\
        & GCN-Align & .561/.262/.357 & .233/.161/.190 & .148/.111/.127 & .113/.086/.098 & .089/.066/.076 & .077/.056/.065 \\
        & AlignE & .757/.394/.518 & .399/.274/.325 & .305/.202/.243 & .245/.154/.189 & .210/.121/.154 & .195/.104/.136 \\
        & AliNet & .653/.361/.465 & .275/.289/.282 & .187/.226/.205 & .144/.180/.160 & .124/.155/.138 & .115/.138/.126 \\
        & KEGCN & .716/.214/.330 & .344/.125/.184 & .260/.090/.134 & .237/.076/.115 & .201/.058/.089 & .169/.045/.071 \\
        & Dual-AMN & .862/.629/.727 & .503/.443/.471 & .394/.331/.359 & .351/.273/.307 & .322/.237/.273 & .313/.214/.254 \\
    \midrule
        \parbox[t]{3mm}{\multirow{3}{*}{\rotatebox[origin=c]{90}{Induct.}}} 
        & $\text{MEAN}^+$ & .840/.585/.690 & .514/.415/.459 & .387/.305/.341 & .314/.235/.269 & .273/.191/.225 & .254/.169/.203 \\
        & $\text{LAN}^+$ & .845/.594/.697 & .506/.410/.453 & .379/.300/.335 & .304/.227/.260 & .269/.188/.222 & .247/.162/.195 \\
        & DINGAL-O & .540/.224/.317 & .381/.165/.231 & .329/.124/.180 & .258/.092/.136 & .247/.073/.112 & .227/.061/.096 \\
    \midrule
        \multicolumn{2}{l|}{\modelname} & \textbf{.866}/\textbf{.634}/\textbf{.732} & \textbf{.569}/\textbf{.520}/\textbf{.543} & \textbf{.453}/\textbf{.421}/\textbf{.436} & \textbf{.387}/\textbf{.351}/\textbf{.369} & \textbf{.351}/\textbf{.301}/\textbf{.324} & \textbf{.325}/\textbf{.265}/\textbf{.292} \\
        \multicolumn{2}{l|}{w/o TA} & NA / NA / NA & .559/.516/.537 &.443/.417/.430 & .379/.348/.363 & .342/.299/.319 & .315/.263/.287 \\
        \multicolumn{2}{l|}{w/o TA \& ASA} & NA / NA / NA & .548/.511/.528 & .431/.413/.421 & .367/.342/.354 & .334/.293/.312 & .311/.256/.281 \\
        \multicolumn{2}{l|}{Retraining} & NA / NA / NA & .516/.437/.473 & .409/.339/.370 & .372/.284/.322 & .348/.247/.289 & .331/.224/.267 \\
    \bottomrule
    \end{tabular}}
\end{table*}

\subsubsection{Ablation study.}
To investigate the impact of each design of \modelname, also to give a fairer comparison between \modelname and baselines, we discard certain parts of \modelname and present three variants as follows:
\begin{itemize}
    \item \modelname w/o TA. In the finetuning process, we discard the selected trustworthy entity alignment and only train on the affected seed alignment. 
    \item \modelname w/o TA \& ASA. We discard both the selected trustworthy alignment and the affected seed alignment. Thus, our method requires no finetuning and reduces to an inductive method. The entity reconstruction method generates embeddings for new entities using their neighbors.
    \item \modelname retraining. Same as the retraining baselines, \modelname treats each snapshot as at $t=0$. Old predicted entity alignment is totally replaced by newly predicted entity alignment rather than being integrated. 
\end{itemize}

We show the results of three variants in Tables~\ref{tab:zh_main_results}, \ref{tab:ja_main_results} and \ref{tab:fr_main_results}.
The variants inherit the trained \modelname at $t=0$ and perform respectively afterwards.
We can notice a performance drop when discarding selected trustworthy alignment.
Bigger declines are seen if further dropping the affected seed alignment. 
This demonstrates the effectiveness of both selected trustworthy alignment and affected seed alignment replay.
For \modelname retraining, though it performs much worse than \modelname, it still outperforms all retraining baselines, including Dual-AMN, 
which indicates the effectiveness of entity reconstruction.

\subsubsection{Discovering new alignment.}
\begin{figure}[t]
\begin{floatrow}
\capbtabbox{
  \resizebox{0.42\textwidth}{!}{
\setlength\tabcolsep{2pt}
\begin{tabular}{cl|ccccc}
\toprule
    & & $t=1$ & $t=2$ & $t=3$ & $t=4$ & $t=5$  \\
\midrule
    \parbox[t]{3mm}{\multirow{5}{*}{\rotatebox[origin=c]{90}{Retraining}}}
    & MTransE & .075 & .055 & .032 & .023 & .013 \\
    & GCN-Align & .049 & .031 & .028 & .014 & .012 \\
    & AlignE & .137 & .099 & .067 & .057 & .040 \\
    & AliNet & .148 & .149 & .118 & .124 & .085 \\
    & KEGCN & .059 & .046 & .026 & .026 & .021 \\
    & Dual-AMN & .204 & .164 & .128 & .113 & .094 \\
\midrule
    \parbox[t]{3mm}{\multirow{3}{*}{\rotatebox[origin=c]{90}{Induct.}}} 
    & $\text{MEAN}^+$ & .170 & .142 & .106 & .098 & .078  \\
    &$\text{LAN}^+$ & .167 & .140 & .109 & .095 & .076 \\
    &DINGAL-O & .003 & .007 & .007 & .008 & .006 \\
\midrule
    \multicolumn{2}{l|}{ContEA} & \textbf{.205} & \textbf{.167} & \textbf{.140} & \textbf{.116} & \textbf{.095} \\
\bottomrule
\end{tabular}
}
}{
  \caption{Recall of the alignment containing new entities on $\text{DBP}_\text{ZH-EN}$.}
  \label{tab:new_entity}
}
\hspace{12pt}
\ffigbox{
  \includegraphics[width=0.46\textwidth]{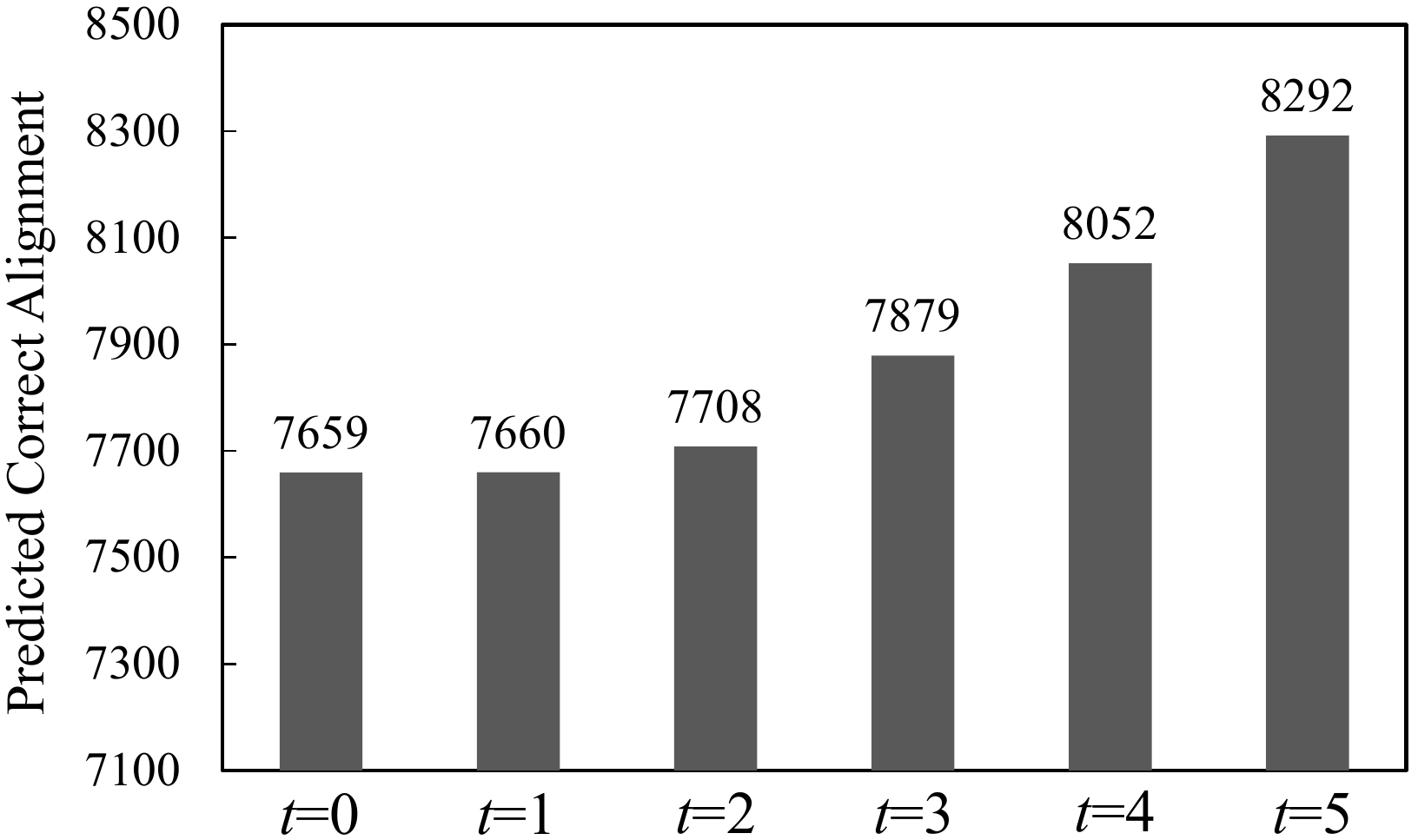}
}{
  \caption{Size growth of predicted correct  alignment on $\text{DBP}_\text{ZH-EN}$.}
  \label{fig:alignment_growth}
}
\end{floatrow}
\end{figure}

Next, we present the performance of \modelname on discovering alignment for new entities. 
At time $t=\{1,2,3,4,5\}$, we collect the final predicted alignment that involves new entities, and calculate the recall value by comparing it with the gold test alignment containing new entities.
We show the results on $\text{DBP}_\text{ZH-EN}$ in Table~\ref{tab:new_entity}.
\modelname reaches the highest recall against all baselines, which indicates the advantage of our method in discovering alignment for new entities. 
We can also notice that the recalls on gold alignment about new entities are significantly lower than those on all gold alignment. 
This is because new entities tend to be sparsely-linked, which hinders the alignment models from matching them correctly.
Also, Fig.~\ref{fig:alignment_growth} illustrates the growth of the total correctly predicted alignment of \modelname.
At time $t$, the size of total correctly predicted alignment is calculated as $|\mathcal{A}^t_{\text{p}}| \times \rm{Recall}$ (R in Table~\ref{tab:zh_main_results}).
The results show that \modelname can find an increasing size of correct entity alignment as KGs grow,
which fulfills the proposal of continual entity alignment.


\subsubsection{Efficiency.}
\begin{figure}[t!]
\centering
\includegraphics[width=\textwidth]{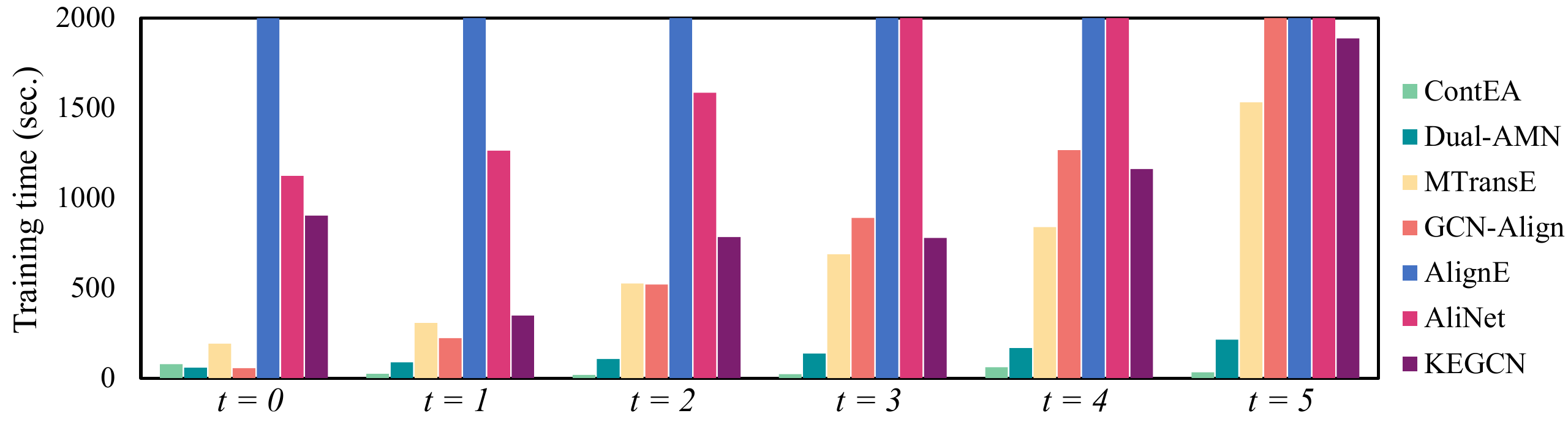}
\caption{Time cost comparison of \modelname and retraining baselines. We report the average time cost on the three datasets.}
\label{fig:time}
\centering
\end{figure}

We compare the training efficiency of \modelname with retraining baselines. 
Note that, since inductive baselines have no training process as new triples come, we do not include them here. 
We run all experiments on a server outfitted with 512GB memory, two Xeon Gold 6326 CPUs, and four RTX A6000 GPUs.
Fig.~\ref{fig:time} depicts the average time cost on three datasets at different snapshots. 
We set the ceiling of vertical axis to 2,000 seconds for better presentation. 
We can see that \modelname has significantly less training time, which shows a part of its superiority in tackling the continual entity alignment task.

\subsection{Further Analysis}

\subsubsection{Incorporating entity names.}
Here we explore the advantage of leveraging entities' names.
Practically, we use \verb|fasttext| library to generate name embeddings for entities. 
Since the original word embedding dimension of \verb|fasttext| is 300, to make the embedding space scalable, we reduce the dimension to 100 using the official dimension reducer.\footnote{\url{https://fasttext.cc/docs/en/crawl-vectors.html}} 
Also, we involve Google Translate\footnote{\url{https://translate.google.com/}} (G.T.) as a competing method. 
For a cross-lingual dataset, we first translate entities from two KGs into the same language (both in English or non-English), 
and then calculate name similarity using Levenshtein distance, a popular measurement in linguistics \cite{LD} and ontology matching \cite{Similarity}. 
The bidirectional nearest neighbor search is also used later to obtain predicted trustworthy entity alignment, 
which are compared with the gold test set to calculate P, R, and F1 scores.
We list the F1 results in Table~\ref{tab:names}. 
By utilizing name attributes, \modelname (\verb|fasttext|) outperforms \modelname with a large margin. 
Google Translate gives satisfactory and robust performance over time, as powerful as expected. 
It performs more stably and is less sensitive to KGs' size, 
with outperforming \modelname (\verb|fasttext|) on most snapshots of the three datasets except the first snapshot. 

\begin{table*}[!tb]\setlength\tabcolsep{2pt}
\centering
\caption{F1 results comparison when incorporating name attribute of entities.}
\label{tab:names}
\resizebox{\textwidth}{!}{
    \begin{tabular}{l|cccccc|cccccc|cccccc}
    \toprule
        & \multicolumn{6}{c|}{$\text{DBP}_\text{ZH-EN}$} & \multicolumn{6}{c|}{$\text{DBP}_\text{JA-EN}$} & \multicolumn{6}{c}{$\text{DBP}_\text{FR-EN}$} \\
        \cmidrule(lr){2-7} \cmidrule(lr){8-13} \cmidrule(lr){14-19} 
        & $t=0$ & $t=1$ & $t=2$ & $t=3$ & $t=4$ & $t=5$ & $t=0$ & $t=1$ & $t=2$ & $t=3$ & $t=4$ & $t=5$ & $t=0$ & $t=1$ & $t=2$ & $t=3$ & $t=4$ & $t=5$ \\
    \midrule
        Google Translate & .550 & .504 & .473 & .451 & .434 & .420 & .677 & .635 & .617 & .595 & .587 & .586 & .749 & .685 & .658 & .645 & .636 & .627 \\
        \modelname (fasttext) & .709 & .556 & .471 & .411 & .367 & .334 & .754 & .615 & .521 & .456 & .409 & .374 & .780 & .613 & .510 & .446 & .404 & .368 \\
        \modelname $\cup$ G.T. & .826  & .645 & .571 & .506 & .466 & .441 & .886 & .742 & .677 & .637 & .612 & .604  & .892  & .740  & .682  & .650 & .633  & .623 \\
    \bottomrule
    \end{tabular}}
\end{table*}

We further explore the combination of \modelname and Google Translate. 
To do so, we combine their predicted alignment when searching the nearest neighbor from one KG to the other, then take a bidirectional intersection to get the final combined predicted alignment. 
The results of this combination are shown in the last row.
We can see that their combination outperforms both Google Translate and \modelname in almost all snapshots of three datasets.
We believe that when Google Translate fails to align an entity, \modelname can be a practical alternative. 

\begin{table*}[!tb]\setlength\tabcolsep{2pt}
\centering
\caption{Case study on previously predicted alignment getting corrected.}
\label{tab:case_study}
\resizebox{\textwidth}{!}{
    \begin{tabular}{c|lc|lc}
    \toprule
        & \multicolumn{2}{c|}{$t=0$} & \multicolumn{2}{c}{$t=5$} \\
        \cmidrule(lr){2-3} \cmidrule(lr){4-5} 
        & Predicted alignment & Sim. & Predicted alignment & Sim. \\
    \midrule
        \multirow{2}{*}{$\text{DBP}_\text{ZH-EN}$} & \textit{(\begin{CJK*}{UTF8}{gbsn}我是歌手\end{CJK*}, Hunan Television)} & .204 & \textit{(\begin{CJK*}{UTF8}{gbsn}湖南卫视\end{CJK*}, Hunan Television)} & .530 \\
        & \textit{(\begin{CJK*}{UTF8}{gbsn}呼和浩特市\end{CJK*}, Baotou)} & .243 & \textit{(\begin{CJK*}{UTF8}{gbsn}包头市\end{CJK*}, Baotou)} & .541 \\
        \midrule
        \multirow{2}{*}{$\text{DBP}_\text{JA-EN}$} & \textit{(\begin{CJK}{UTF8}{min}スウェーデン語\end{CJK}, Finnish language)} & .210 & \textit{(\begin{CJK}{UTF8}{min}フィンランド語\end{CJK}, Finnish language)} & .316 \\
        & \textit{(\begin{CJK}{UTF8}{min}フランス人\end{CJK}, Spaniards)} & .265 & \textit{(\begin{CJK}{UTF8}{min}フランス人\end{CJK}, French people)} & .367 \\
        \midrule
        \multirow{2}{*}{$\text{DBP}_\text{FR-EN}$} & \textit{(Révolution française, Réunion)} & .262 & \textit{(La Réunion, Réunion)} & .403 \\
        & \textit{(Stade de Wembley, White Hart Lane)} & .302 & \textit{(Stade de Wembley, Wembley Stadium)} & .380 \\
        \bottomrule
    \end{tabular}}
\end{table*}

\subsubsection{Case study on correcting previous alignment.}
Last, we present several cases in Table~\ref{tab:case_study} about the previously predicted alignment getting corrected in later finetuning processes. 
We save the predicted alignment and their similarity scores at time $t=0$ and $t=5$, and juxtapose two alignment pairs from each time that involve the same entity. 
We manually check the list of juxtaposition and notice that the predicted alignment at $t=0$ is usually incorrect with smaller similarity scores, while their counterparts at $t=5$ are correct with higher similarity scores. 
This indicates the ability of \modelname on self-correction. 
Meanwhile, the two entities in falsely predicted alignment at $t=0$ are not totally irrelevant. For example, in the second case from the $\text{DBP}_\text{FR-EN}$ dataset, both \textit{Stade de Wembley} and \textit{White Hart Lane} are Stadiums in London. 
In the first case from the $\text{DBP}_\text{ZH-EN}$ dataset, \textit{\begin{CJK*}{UTF8}{gbsn}我是歌手\end{CJK*}} is a popular TV show made by \textit{Hunan Television}. And in the first case from the $\text{DBP}_\text{JA-EN}$ dataset, \begin{CJK}{UTF8}{min}スウェーデン語\end{CJK} means Swedish language (Sweden and Finland are two neighboring Nordic countries). 
This gives an interesting insight on how \modelname predicts entity alignment with slight inaccuracy. 

\section{Related Work}

\medskip\noindent\textbf{Static entity alignment.}
Most existing embedding-based entity alignment methods focus on static KGs. 
They can usually be classified into two categories regarding the techniques of their KG encoders: translation-based \cite{MTransE,AKE,OTEA,BootEA,TransEdge,IPTransE} and GNN-based \cite{Dual-AMN,MRAEA,AliNet,GCN-Align,GMNN,RDGCN,NMN}. 
The former family adopts translation-based KG embedding (KGE) techniques \cite{TransE,TransH} to embed entities, and map cross-graph entities into a unified space based on pre-aligned entity pairs.
The encoder of GNN-based entity alignment methods learns a shared neighborhood aggregator to embed entities in different KGs.
They have gained overwhelming popularity in recent years due to their strong ability to capture the structural information using a subgraph around an entity, rather than a single triple.
For more details, there are several surveys \cite{OpenEA,EA_survey_AIOpen} that comprehensively summarize the recent advances.

\medskip\noindent\textbf{Dynamic entity alignment.}
As far as we know, DINGAL \cite{DINGAL} is the only entity alignment method that addresses the dynamics of KGs. 
In its dynamic scenario, new triples are added into KGs as well as new pre-known alignment provided along with these new entities. 
A variant of DINGAL, named DINGAL-O, is also proposed in their work to handle a similar setting as ours where the pre-known alignment does not grow.
DINGAL-O is an inductive method that leverages prior-learned model parameters to predict new alignment.
Particularly, they use name attributes to generate word embeddings for entity initialization. 

\medskip\noindent\textbf{Inductive knowledge graph embedding.}
The study on dynamic KG embedding has drawn lots of attention over the years.
Powered by GNN, many inductive embedding methods for KG completion are proposed to generate embeddings for new entities. 
Early inductive methods either focus on semi-inductive settings where new entities are connected to existing KG and making inferences between new entities and existing entities \cite{MEAN,VNNetwork,LAN,DKRL}, or fully-inductive settings where new entities form independent graphs and making inferences among new entities \cite{TACT,GraIL}. 
Later inductive method \cite{INDIGO} intend to tackle both settings.
Meanwhile, some inductive KG embedding methods focus on special tasks like few-shot learning \cite{DKRL} and hyper-relational KG completion \cite{Hyper-relational}.
Specifically, as the first inductive KG embedding method, MEAN \cite{MEAN} learns to represent entities using their neighbors by simply mean-pooling the information of neighboring entity-relation pairs.
LAN \cite{LAN} advances MEAN by incorporating a rule-based attention and a GNN-based attention on entity-relation pairs in the pooling process. 

\section{Conclusion and Future Work}
In this paper, considering the growth nature of real-world KGs, we focus on an entity alignment scenario where both graphs are growing, and address a new task named continual entity alignment.
We propose a novel method \modelname as a solution to the task. 
Also, we construct three datasets to imitate the scenario and conduct extensive experiments.
The experimental results show the superiority of \modelname in terms of effectiveness and efficiency against a list of retraining and inductive baselines.  
For future work, there are many promising improvements and extensions to the current proposal.
Regarding the setting, future studies can consider more complex scenarios such as the addition of new relations, the addition of new pre-known alignment, and even the deletion of entities and triples.
As to the method, more reliable and comprehensive trustworthy alignment update strategies are necessary to handle intricate alignment conflict cases.

\paragraph*{Supplemental Material Statement:} The source code, detailed hyperparameters, and constructed datasets are available at our GitHub repository.\footnote{\url{https://github.com/nju-websoft/ContEA}}

\bigskip
\noindent\textbf{Acknowledgments.} 
This work was supported by National Natural Science Foundation of China (No. 61872172), 
Beijing Academy of Artificial Intelligence (BAAI),
and Collaborative Innovation Center of Novel Software Technology \& Industrialization. 
Zequn Sun was also grateful for the support of Program A for Outstanding PhD Candidates of Nanjing University. 

\bibliographystyle{splncs04}
\bibliography{reference}
\end{document}